\DocumentMetadata{
  pdfversion=2.0,pdfstandard=ua-2,
  testphase={phase-III,firstaid,math,title}
}

\documentclass[sigconf, screen]{acmart-tagged}

\AtBeginDocument{%
  }

\usepackage{microtype}

\setcopyright{cc}
\setcctype{by}
\acmDOI{10.1145/3776734.3794357}
\acmYear{2026}
\copyrightyear{2026}
\acmISBN{979-8-4007-2321-6/2026/03}
\acmConference[HRI Companion '26]{Companion Proceedings of the 21st ACM/IEEE International Conference on Human-Robot Interaction}{March 16--19, 2026}{Edinburgh, Scotland, UK}
\acmBooktitle{Companion Proceedings of the 21st ACM/IEEE International Conference on Human-Robot Interaction (HRI Companion '26), March 16--19, 2026, Edinburgh, Scotland, UK}
\received{2025-12-08}
\received[accepted]{2026-01-12}

\begin{document}

\title[Vision-Language System using Open-Source LLMs for Gestures in Medical Interpreter Robots]{Vision-Language System using Open-Source LLMs for Consent and Instruction Gestures in Medical Interpreter Robots}

\author{Thanh-Tung Ngo}
\orcid{0009-0004-0065-8600}
\affiliation{%
  \institution{Technological University Dublin}
  \city{Dublin}
  \country{Ireland}
}
\email{thanhtung.ngo@tudublin.ie}

\author{Emma Murphy}
\orcid{0000-0001-6738-3067}
\affiliation{%
  \institution{Technological University Dublin}
  \city{Dublin}
  \country{Ireland}
}
\email{emma.x.murphy@tudublin.ie}

\author{Robert J. Ross}
\orcid{0000-0001-7088-273X}
\affiliation{%
  \institution{Technological University Dublin}
  \city{Dublin}
  \country{Ireland}
}
\email{robert.ross@tudublin.ie}

\begin{abstract}

Effective communication is vital in healthcare, especially across language barriers, where non-verbal cues and gestures are critical. 
This paper presents a privacy-preserving vision-language framework for medical interpreter robots that detects specific speech acts (consent and instruction) and generates corresponding robotic gestures.
Built on locally deployed open-source models, the system utilizes a Large Language Model (LLM) with few-shot prompting for intent detection.
We also introduce a novel dataset of clinical conversations annotated for speech acts and paired with gesture clips.
Our identification module achieved 0.90 accuracy, 0.93 weighted precision, and a 0.91 weighted F1-Score.
Our approach significantly improves computational efficiency and, in user studies, outperforms the speech-gesture generation baseline in human-likeness while maintaining comparable appropriateness.
  
\end{abstract}





\maketitle

\section{Introduction}

Gestures play a crucial role in healthcare communication, particularly when language barriers exist between healthcare providers and patients. 
Physicians globally use similar gestures when discussing medical conditions, reflecting the universal nature of biomedical knowledge and its embodied expression across cultures \cite{quasinowski_hearts_2023}. 
In interpreted healthcare encounters, gestures serve important semantic functions, with approximately 70\% of doctors' and patients' body-oriented gestures providing information not conveyed in speech \cite{gerwing_body-oriented_2019}. 
In pain communication specifically, gestures serve three distinct functions: pointing, iconic, and symbolic, with multiple communicative modalities interwoven in patient accounts \cite{hyden_pain_2002}. 
Understanding gestural communication becomes particularly important given increasing language barriers between practitioners and patients in diverse healthcare settings \cite{meuter_overcoming_2015}.

Nonverbal communication is critical in healthcare for validating consent and conveying instructions. Consent-related gestures reinforce verbal assurances and bridge language gaps, ensuring genuine patient autonomy. Simultaneously, instructional gestures transform abstract medical concepts into concrete visual demonstrations \cite{10.1093/oxfordhb/9780199795833.013.021}, significantly reducing cognitive load. By integrating these nonverbal cues, clinicians improve treatment adherence and minimize safety risks associated with miscommunication.

Technological support for medical interpretation has evolved significantly, ranging from mobile video services \cite{ji_utility_2021} to direct translation software. While these tools reduce wait times and costs, they face adoption barriers due to accuracy concerns, particularly in medico-legal contexts \cite{ji_utility_2021, ogunnaike_prevalent_2022}. 
Specialized systems like the VIP (Voice-text Integrated system for interPreters) introduce technology-based terminology workflows for simultaneous interpretation \cite{corpas_pastor_technology_2022}.
These existing solutions are unable to convey nonverbal communication. Robotic systems, capable of integrating verbal and gestural cues, present a novel opportunity to bridge this gap.

A significant gap remains in deploying robots capable of informative gestures in healthcare. This challenge is threefold. First, the scarcity of domain-specific datasets limits the ability to learn contextually appropriate medical movements. Second, existing generation techniques focus on general conversation, failing to capture the precision required for medical instructions or account for cultural and individual gesture variability. Third, the limited computational resources of robotic platforms constrain the real-time execution of sophisticated models. These technical hurdles, combined with the strict reliability standards of high-stakes clinical environments, necessitate a framework that bridges the gap between technological capability and real-world application.
Our contributions are:
\begin{itemize}
    \item We introduce a clinical conversation dataset featuring videos and transcripts with sentence-level gesture annotations.
    \item We propose a privacy-preserving, lightweight LLM-based gesture sentence detector optimized for local computation, ensuring both speed and security.
    \item We present a pipeline that maps human pose kinematics to robot motor commands, enabling human-like gestures.
    \item We implement a framework integrating the detector and the pipeline, evaluating its performance on the Pepper robot.
\end{itemize}

\section{Related Work}

\subsection{Gesture Detection in Healthcare}

Recent research has explored various approaches to detecting and recognizing relevant gestures in healthcare.
Deep Learning and IoT systems recognize medically relevant gestures, helping patients communicate symptoms like fever or pain and overcoming language barriers \cite{b_iot-enhanced_2024}.
Convolutional Neural Networks (CNNs) enable patient monitoring and allow individuals, particularly bedridden or paralyzed patients unable to use conventional methods, to send urgent messages \cite{rahul_hand_2024, k_assistance_2023}.
Technically, the gesture recognition process involves hand and finger detection with background subtraction and image processing \cite{k_assistance_2023}.
Finally, the technology supports multidisciplinary collaborations advancing automatic clinical behavior analysis \cite{hammal_face_2020}.

\subsection{Speech-Gesture Generation}

Data-driven gesture generation has evolved from early statistical analysis \cite{10.1145/1833349.1778861} to advanced deep learning. Initial deep learning approaches used Multi-Layer Perceptrons (MLPs) \cite{10.1145/3382507.3418815} for realistic, context-aware gestures, and CNNs \cite{10.1145/3472306.3478335} to capture spatial relationships. Subsequently, Recurrent Neural Networks   \citep{yoon_robots_2019,liu_learning_2022} improved sequential alignment with human motor behavior, while Transformers \cite{transformer} leveraged long-range dependencies for coherence. Most recently, generative models including VQ-VAEs \cite{yang_qpgesture_2023}, GANs \cite{liu_speech-gesture_2023}, and diffusion models \cite{zhao_diffugesture_2023, tonoli_gesture_2023, ijcai2023p650} have further enhanced synchronization and diversity, closely matching human performance.

\section{System Overview}

\begin{figure}[!h]
  \centering
  \includegraphics[width=\columnwidth]{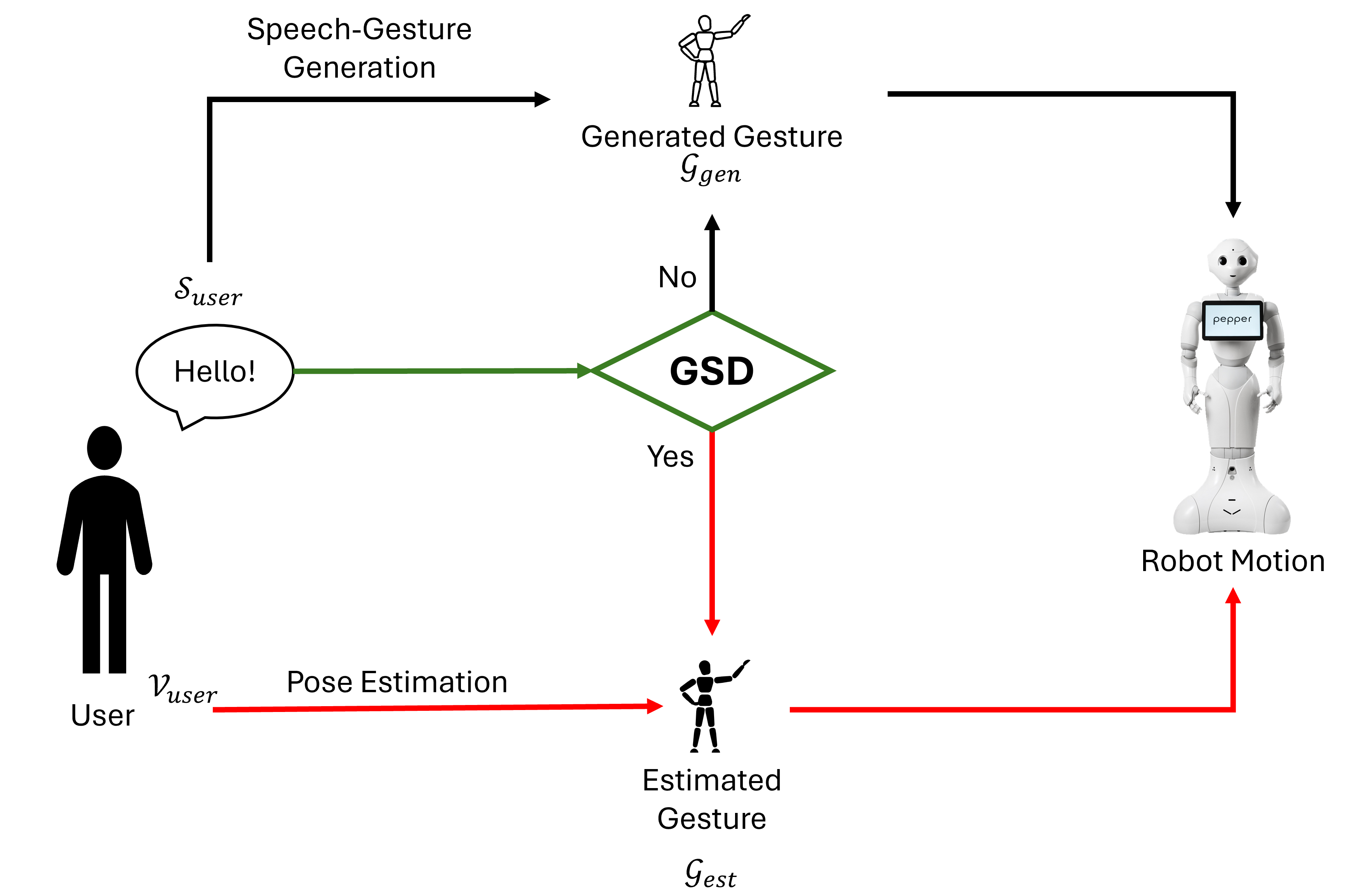}
  \caption{Overview of our proposed vision-language system for gesturing medical interpreter robots.}
  \Description{A flowchart illustrating the system architecture. A 'User' provides two inputs: speech containing the text 'Hello!', and visual input. The speech input enters a decision diamond labeled 'GSD'. If the GSD output is 'No' (black path), the system proceeds to 'Speech-Gesture Generation' to create a 'Generated Gesture'. If the GSD output is 'Yes' (red path), the system performs 'Pose Estimation' on the user's visual input to create an 'Estimated Gesture'. Both paths converge to drive the final 'Robot Motion' on a Pepper robot.}
  \label{fig:System-Overview}
\end{figure}

Figure \ref{fig:System-Overview} shows an overview of our framework. 
The system captures user speech ($\mathcal{S}_{user}$) from the robot’s microphone and video ($\mathcal{V}_{user}$) from its camera. 
A Gesture Sentence Detection (GSD) module, implemented with an LLM, classifies each utterance as \textit{Consent}, \textit{Instruction}, or \textit{Neither}. 
If classified as \textit{Consent} or \textit{Instruction}, the Human-Mimic module uses ($\mathcal{V}_{user}$) to reproduce the user’s gesture, producing ($\mathcal{G}_{est} \in \mathbb{R}^{n_{est} \times 12}$), the Pepper robot’s 12 joint angles over ($n_{est}$) timestamps. Otherwise, the Speech-Gesture Generation module generates an appropriate gesture ($\mathcal{G}_{gen} \in \mathbb{R}^{n_{gen} \times 12}$).

To protect user privacy in healthcare settings, all models are open-source and run locally, ensuring no data leaves the device. The system and experiments were executed on a PC with an NVIDIA RTX 4090 GPU, AMD Ryzen 9 7950X CPU, and Ubuntu 20.04.

\section{Gesture Sentence Detection}

\subsection{Dataset} \label{subsec:dataset}

\begin{figure}[!h]
  \centering
  \includegraphics[width=\linewidth]{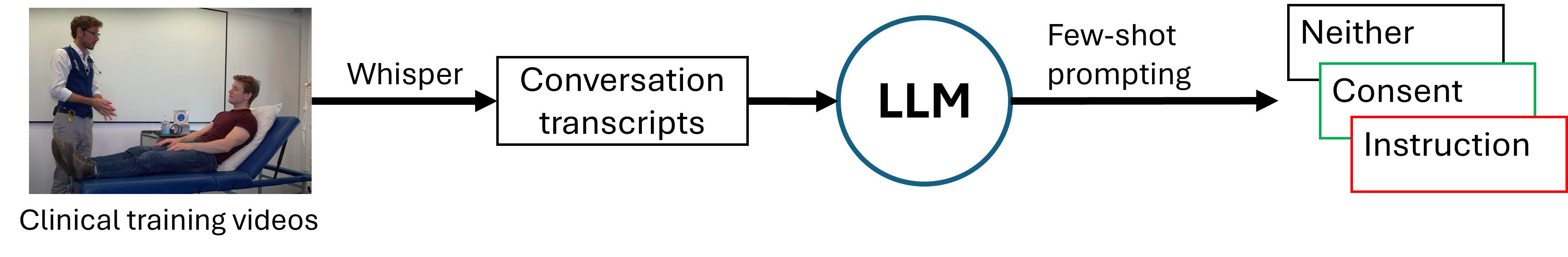}
  \caption{Pipeline for using an LLM to create the clinical conversation dataset.}
  \Description{A data processing pipeline diagram starting with 'Clinical training videos' showing a doctor and patient. The video audio is processed by 'Whisper' to produce 'Conversation transcripts'. These transcripts are input into an 'LLM' using 'Few-shot prompting' to generate classification labels. The final outputs are three categories labeled 'Neither,' 'Consent,' and 'Instruction'.}
  \label{fig:sentence-dataset}
\end{figure}

Figure \ref{fig:sentence-dataset} illustrates the creation process for our clinical conversation dataset. 
58 clinical training videos were selected from the public YouTube channel \textit{Dr James Gill}\footnote{https://www.youtube.com/@DrJamesGill}. 
The videos were transcribed using the \textit{OpenAI Whisper} \cite{10.5555/3618408.3619590} model. 
Because Whisper often split long utterances, we implemented a script to reconstruct full sentences and align them with their video timestamps, yielding 3,736 complete sentences. 
The 58 source videos were then segmented into 3,736 short clips, each corresponding to one reconstructed sentence.

Sentences were classified into \textit{Consent}, \textit{Instruction}, or \textit{Neither} using \textit{gpt-oss:20b} \cite{openai2025gptoss120bgptoss20bmodel}, \textit{qwen3:30b} \cite{yang2025qwen3technicalreport}, and deepseek-r1:8b \cite{guo_deepseek-r1_2025}, yielding a 21.95\% inter-model disagreement rate. 
A human annotator resolved these conflicts, validated all Consent/Instruction outputs, and audited the top 20\% longest \textit{Neither} samples. 
This validation process confirmed a 92\% agreement rate between the human and the model consensus. 
The final dataset contains 3,736 labeled sentences (117 Consent, 912 Instruction, 2,707 Neither).
The annotation structure is shown below:

\begin{verbatim}
    Sentence: "\{sentence\}"
    Classification: [INSTRUCTION/CONSENT/NEITHER]
    Reasoning: [Brief explanation]
\end{verbatim}

\subsection{Gesture Sentence Detection} \label{subsec:GSD-method}

\begin{figure}[!h]
  \centering
  \includegraphics[width=\linewidth]{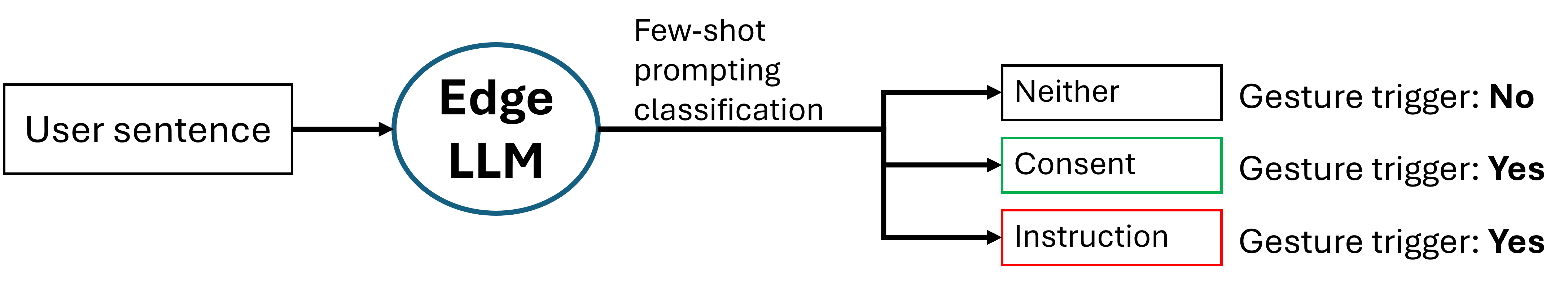}
  \caption{The on-device gesture sentence detection module.}
  \Description{A classification flowchart starting with a 'User sentence' input. The text is processed by an 'Edge LLM' via 'Few-shot prompting classification'. The system outputs one of three labels: 'Neither' results in 'Gesture trigger: No', while both 'Consent' and 'Instruction' result in 'Gesture trigger: Yes'.}
  \label{fig:sentence-detection}
\end{figure}

The Gesture Sentence Detection workflow (Figure \ref{fig:sentence-detection}) employs a lightweight edge LLM for low-latency, on-device classification into \textit{Consent}, \textit{Instruction}, or \textit{Neither}. We utilized an 11-shot prompting strategy comprising four examples for \textit{Instruction}, four for \textit{Consent}, and three for \textit{Neither}. To enhance generalization, the prompt combines six examples from the curated dataset (Section \ref{subsec:dataset}) with five manually created samples.

Parameters were configured for deterministic quality: we set temperature to 0.1 to minimize randomness, while top-p (0.9) and top-k (10) were used to balance diversity and constrain token selection. To prevent premature truncation of verbose models that generate explanatory text, $num\_predict$ was extended to 1,000 tokens. Additionally, we set the context window ($num\_ctx$) to 10,000 to handle complex prompts and disabled streaming (stream=False) to simplify post-processing by ensuring complete responses.

\section{Robot Gesture Generation}

\subsection{Human-Mimic}

\begin{figure}[!h]
  \centering
  \includegraphics[width=\linewidth]{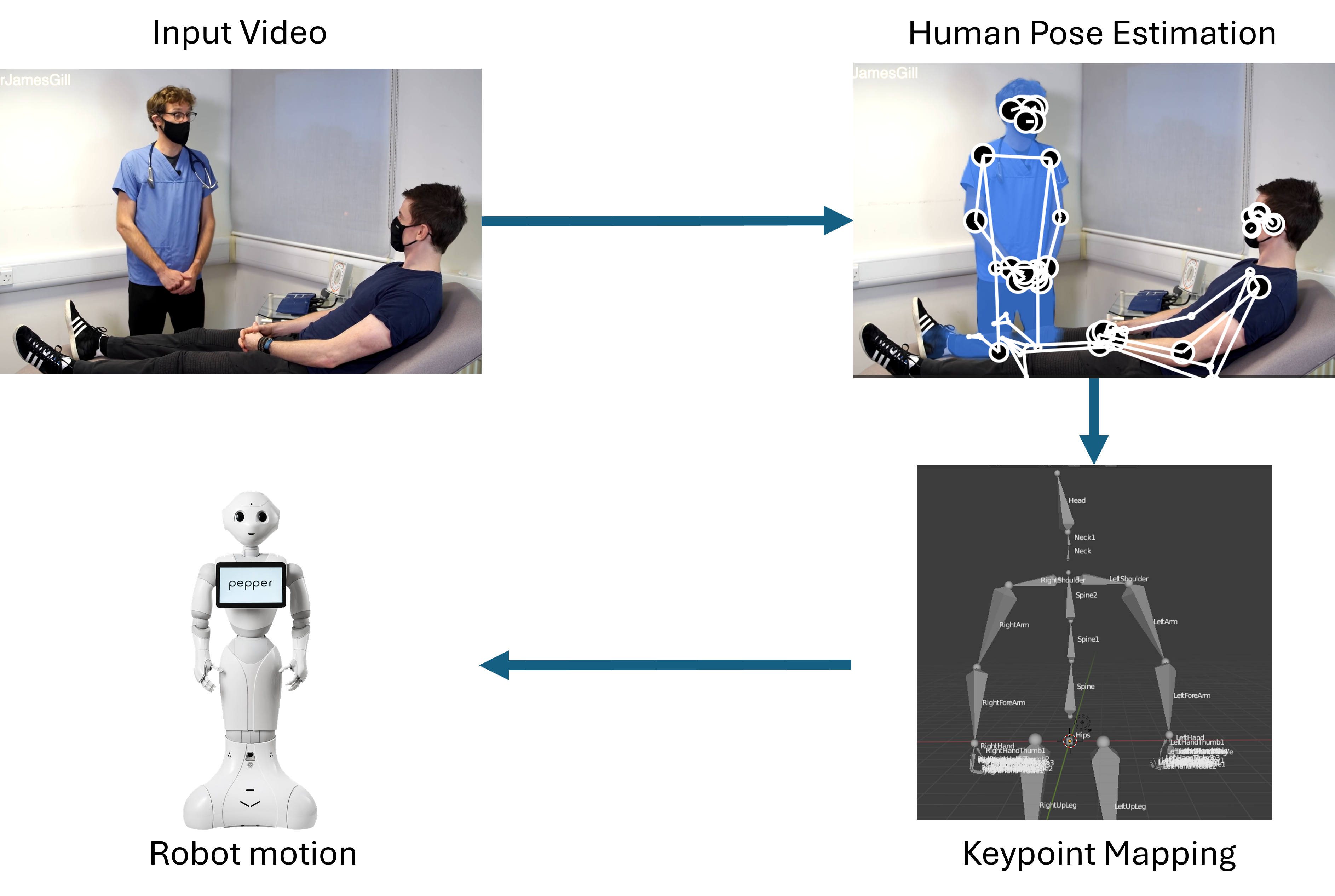}
  \caption{Conversation videos to robot motions workflow.}
  \Description{A four-stage pipeline diagram illustrating human-to-robot motion transfer. The process begins with an Input Video of two people, flows to Human Pose Estimation, where skeletal keypoints are detected on the subjects, proceeds to Keypoint Mapping, showing a 3D skeletal rig with labeled joints (e.g., Head, Spine, Arms), and concludes with Robot motion displayed on a Pepper humanoid robot.}
  \label{fig:human-robot}
\end{figure}

Figure \ref{fig:human-robot} outlines our pipeline for translating a human's gesture from a video into the corresponding robot's motion.
The process begins by using a pose estimation model to estimate the coordinates of key skeletal landmarks (e.g., shoulders, elbows, and neck) from the user's video. 
These coordinates are then fed into a keypoint mapping function, translating the human's pose into an array of joint angles trajectory $\mathcal{G}_{est} \in \mathbb{R}^{n_{est} \times 12}$ for the Pepper robot. 
Finally, the array $\mathcal{G}_{est}$ is converted to actuator commands by the \textit{NAOqi Python SDK} \footnote{http://doc.aldebaran.com/2-5/dev/python/index.html}.
The Human-Mimic module was developed based on the \textit{pepper-blazepose} \cite{khalil_human_2021} package.
To adhere to the speed limitations of the robot's joint angles, we set the scaling factor to 12, compensating for the characteristics of the dataset (section \ref{subsec:dataset}).

For accurate and resource-efficient pose estimation, \textit{MediaPipe Pose Landmarker}\footnote{https://ai.google.dev/edge/mediapipe/solutions/vision/pose\_landmarker} and \textit{YOLO11-pose}\footnote{https://docs.ultralytics.com/tasks/pose/} are the most suitable candidates. We ultimately selected MediaPipe because its CPU optimization aligns with the computational constraints of robotic hardware. Furthermore, MediaPipe’s detect-and-track pipeline produces less temporal jitter than YOLO11-pose's single-frame prediction method. This stability is critical for our application, directly translating to smoother robot motion trajectories.

\subsection{Speech-Gesture Generation}

We employ the Semantic Gesticulator (SG) \cite{zhang_semantic_2024} to generate semantics-aware co-speech gestures from audio input, outputting motion data in Biovision Hierarchy (BVH) format. To retarget this motion to the Pepper robot, we implemented a pipeline that parses the BVH skeleton and maps specific joints to Pepper’s kinematic structure. For each frame, rotation values are transformed from Euler angles into Pepper’s joint space ($G_{gen} \in \mathbb{R}^{n_{gen} \times 12}$), applied with necessary linear adjustments, and clamped to safe operational limits. However, the high frequency of the source motion (60 Hz) resulted in joint velocities exceeding the robot's safety thresholds. To resolve this, we down-sampled the keyframe sequence by a factor of $N=12$. This approach preserves the total motion duration while effectively reducing commanded velocities, ensuring smooth execution within the robot's mechanical constraints.

\section{Experiments and Results}

\subsection{Gesture Sentence Detection} \label{subsec:GSD-results}

We evaluated nine state-of-the-art lightweight LLMs, selected to represent a range of deployment targets: four medium-sized (7B–8B parameters) LLMs for PCs with GPUs, two small (3B–4B) LLMs for laptops or embedded computers, and three tiny LLMs designed for edge devices.
Following the pipeline in Section \ref{subsec:GSD-method}, each model was evaluated on the dataset described in Section \ref{subsec:dataset}.
A uniform prompt was used across all experiments to ensure a controlled comparison.

\begin{table}[ht!]
\centering
\caption{Evaluation of LLMs on gesture sentence detection}
\label{tab:model_comparison}
\resizebox{\columnwidth}{!}{%
\begin{tabular}{lcccc}
\toprule
\textbf{Model} & \textbf{Accuracy} $\uparrow$ & \textbf{Precision} $\uparrow$ & \textbf{F1-Score} $\uparrow$ & \textbf{Memory (GB)} $\downarrow$ \\
\midrule
qwen3:8b & \textbf{0.90}&\textbf{0.93}& \textbf{0.91}& 7.2\\
 deepseek-r1:8b & 0.83& 0.87& 0.84& 7.2\\
llama3.1:8b& 0.67& 0.83& 0.69& 6.8\\
 mistral:7b& 0.70& 0.83& 0.72& \textbf{6.5}\\
  & & & & \\
 gemma3:4b & \textbf{0.63}& 0.81& \textbf{0.69}& 5.2\\
 llama3.2:3b& 0.56& \textbf{0.83}& 0.60
& \textbf{4.0}\\
  & & & & \\
 qwen3:0.6b& \textbf{0.70}& 0.73& \textbf{0.71}& 2.3\\
 deepseek-r1:1.5b& 0.29& \textbf{0.74}& 0.42& 2.0\\
gemma3:270m& 0.11& 0.07& 0.08& \textbf{0.9}\\ \bottomrule
\end{tabular}
}
\end{table}

\begin{figure}[!h]
  \centering
  \includegraphics[width=\linewidth]{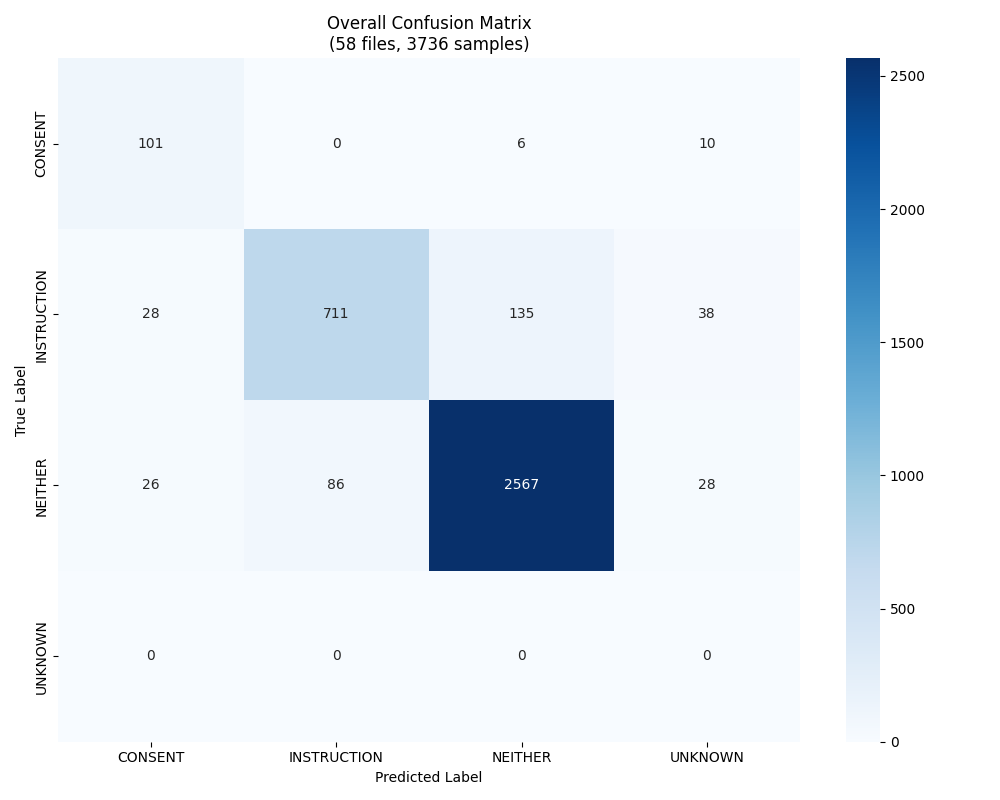}
  \caption{Confusion matrix illustrating the performance of \textit{qwen3:8b}.}
  \Description{A confusion matrix with the following values. True Consent row: 101 predicted Consent, 0 Instruction, 6 Neither, 10 Unknown. True Instruction row: 28 predicted Consent, 711 Instruction, 135 Neither, 38 Unknown. True Neither row: 26 predicted Consent, 86 Instruction, 2567 Neither, 28 Unknown. The matrix uses a blue color scale where darker blue indicates higher frequencies, peaking at 2567.}
  \label{fig:confusion-matrix-qwen3}
\end{figure}

The results of the experiments are shown in Table \ref{tab:model_comparison}.
The classification performance of different LLMs shows clear distinctions between model sizes.
\textit{qwen3:8b}
\cite{yang2025qwen3technicalreport} not only produces highly reliable classifications but also balances both recall and precision effectively, making it the strongest candidate for high-accuracy tasks.
Figure \ref{fig:confusion-matrix-qwen3} shows the confusion matrix for \textit{qwen3:8b}.
Close behind, \textit{deepseek-r1:8b} 
\cite{guo_deepseek-r1_2025} gets slightly lower performance compared to \textit{qwen3:8b} while consuming the same GPU memory.
\textit{llama3.1:8b} \cite{sam_llama_2024} and \textit{mistral:7b} \cite{jiang2023mistral7b} perform moderately well.
Despite their reduced memory demands compared to \textit{qwen3:8b} and \textit{deepseek-r1:8b}, these models show a trade-off where efficiency in resources comes at the expense of predictive reliability.
Likewise, small LLMs like \textit{gemma3:4b} \cite{gemmateam2025gemma3technicalreport} and \textit{llama3.2:3b} \cite{grattafiori2024llama3herdmodels} further illustrate this trend: although their precision remains relatively high, their accuracy and F1-Scores show substantial drops, reflecting difficulty in generalization despite correctly identifying positive cases. 
\textit{qwen3:0.6b} \cite{yang2025qwen3technicalreport} achieves a surprisingly competitive accuracy of 0.70 and F1-Score of 0.71 with just 2.3 GB memory usage, highlighting its efficiency compared to mid-sized models.
In contrast, \textit{deepseek-r1:1.5b} \cite{guo_deepseek-r1_2025} demonstrates a severe imbalance: although its precision is relatively high, its accuracy and F1-Score collapse, indicating that it produces correct predictions very rarely but with confidence when it does. 
The smallest model, \textit{gemma3:270m} \cite{gemmateam2025gemma3technicalreport}, gets very low results, despite being the most memory-efficient option at just 0.9 GB. 
This suggests that extreme reductions in model size can undermine classification capabilities beyond practical thresholds.
The models also generated classes outside the three predefined categories, a behavior indicative of the hallucination phenomenon in LLMs.

\subsection{Robot Gesture Evaluation}

\begin{figure}[!h]
  \centering
  \includegraphics[scale=0.5]{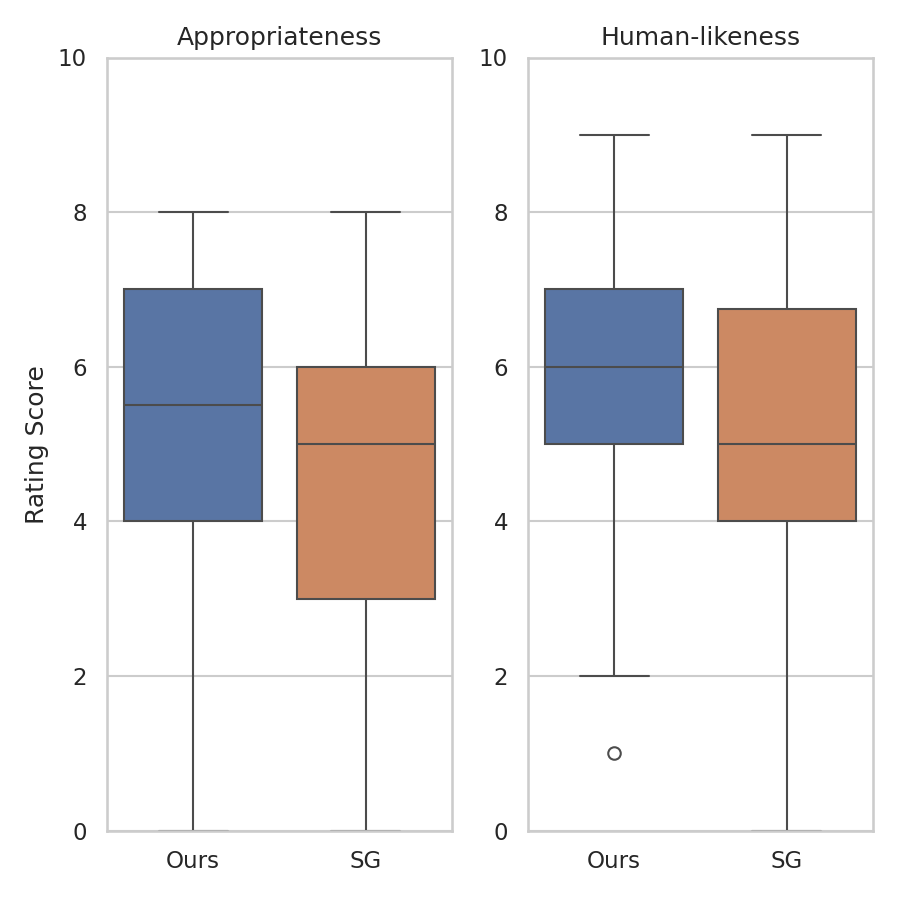}
  \caption{Boxplots visualizing the rating distributions.}
  \Description{Side-by-side boxplots with a y-axis 'Rating Score' ranging from 0 to 10. Appropriateness: 'Ours' (Median 5.50, Q1 4.00, Q3 7.00, Max 8.00, Min 0.00); 'SG' (Median 5.00, Q1 3.00, Q3 6.00, Max 8.00, Min 0.00). Human-likeness: 'Ours' (Median 6.00, Q1 5.00, Q3 7.00, Max 9.00, Min 1.00, with an outlier at 1.00); 'SG' (Median 5.00, Q1 4.00, Q3 6.75, Max 9.00, Min 0.00).}
  \label{fig:boxplot}
\end{figure}

To evaluate the impact of the Human-Mimic module, triggered when GSD identifies \textit{Consent} or \textit{Instruction}, we conducted a within-subject user study (26 participants, aged 25-35) benchmarking our approach against the Semantic Gesticulator (SG) \cite{zhang_semantic_2024}.
Using five random samples from Section \ref{subsec:dataset} (one $\approx$20s and four $\approx$5s clips), we evaluated gesture human-likeness and speech appropriateness following GENEA 2023 Challenge \cite{kucherenko_genea_2023}, incorporating attention checks for validity. 
The comparison results are shown in Figure \ref{fig:boxplot}.
Additionally, we benchmarked computational resource requirements to assess implementation viability; results are presented in Table 2.

\begin{table}[ht!]
\centering
\caption{Comparison of human-likeness, appropriateness and memory usage between our approach and SG (mean values).}
\label{tab:gesture-generation-comparison}
\resizebox{\columnwidth}{!}{
\begin{tabular}{lccc}
\hline
\textbf{Approach} & \textbf{Human-likeness} $\uparrow$ & \textbf{Appropriateness} $\uparrow$ & \textbf{GPU RAM (MB)} $\downarrow$ \\ \hline
SG \cite{zhang_semantic_2024} & 5.24 & 4.76 & 2260 \\
Ours & \textbf{5.78} & \textbf{5.20} & \textbf{3} \\ \hline
\end{tabular}
}
\end{table}

\subsubsection{Human-likeness}
This criterion evaluates robot motion resemblance to humans, independent of speech.
Participants watched silent video clips of both approaches and rated them based on the question: "How human-like is the robot’s gesture in this video?"
Ratings were provided on an 11-point scale from 0 (very bad) to 10 (excellent).
A one-tailed paired sample T-test confirmed that our approach ($M=5.78$) scored significantly higher than the SG ($M=5.24$; $p=0.019$).

\subsubsection{Appropriateness}

This criterion measures gesture-speech alignment, controlling for overall human-likeness. 
Participants watched video clips with speech and rated them based on the question: "How well do the robot’s movements in this video reflect the speech?" 
Responses were collected on the same 0 (very bad) to 10 (excellent) scale.
 A paired-samples T-test yielded a mean difference of 0.44 between our approach ($M = 5.20$) and SG ($M = 4.76$), which was not statistically significant ($p = 0.277$). 
 The 95\% CI $[-0.39, 1.28]$ confirms that the observed numerical preference does not constitute statistical superiority. 

\section{Conclusions}

This paper presented a privacy-preserving vision-language framework that leverages lightweight, on-device LLMs to detect consent and instruction gestures in healthcare conversations with high accuracy.
The user study demonstrated that our approach achieves significantly higher human-likeness while maintaining appropriateness comparable to the speech-gesture generation baseline.
Crucially, the system operates locally with minimal GPU usage, ensuring the data privacy and security essential for clinical settings.
This work, along with the release of our clinical gesture dataset, advances the capability for natural and secure human–robot interaction.

\begin{acks}
This research was conducted with the financial support of Research Ireland under Grant Agreement No. 13/RC/2106\_P2 at ADAPT, the SFI Research Centre for AI-Driven Digital Content Technology.
\end{acks}

\bibliographystyle{ACM-Reference-Format}

\bibliography{gesture}



\end{document}